\documentclass{article}
\usepackage{spconf,amsmath,graphicx}

\usepackage{booktabs, url}
\usepackage{caption} 
\usepackage{hyperref}
\usepackage[style=ieee, maxcitenames=3]{biblatex}
\usepackage{paralist}
\usepackage{todonotes} 
\usepackage{comment}

\title{SURROGATE GRADIENT SPIKING NEURAL NETWORKS AS ENCODERS FOR LARGE VOCABULARY CONTINUOUS SPEECH RECOGNITION}

\name{Alexandre Bittar$^{\,1,2}$, Philip N. Garner$^{\,1}$\thanks{This project received funding under NAST: Neural Architectures for Speech Technology, Swiss National Science Foundation, grant number \href{https://data.snf.ch/grants/grant/185010}{185010}.}}
\address{
    $^1$Idiap Research Institute, Martigny, Switzerland \\
    $^2$\'Ecole Polytechnique Fédérale de Lausanne, Switzerland
    }

\begin{document}
\maketitle
%
\begin{abstract}
    Compared to conventional artificial neurons that produce dense and real-valued responses, biologically-inspired spiking neurons transmit sparse and binary information, which can also lead to energy-efficient implementations. Recent research has shown that spiking neural networks can be trained like standard recurrent neural networks using the surrogate gradient method. They have shown promising results on speech command recognition tasks. Using the same technique, we show that they are scalable to large vocabulary continuous speech recognition, where they are capable of replacing LSTMs in the encoder with only minor loss of performance. This suggests that they may be applicable to more involved sequence-to-sequence tasks.  Moreover, in contrast to their recurrent non-spiking counterparts, they show robustness to exploding gradient problems without the need to use gates.
\end{abstract}

\section{Introduction}
\label{sec:intro}

Artificial Neural Networks (ANNs) have become ubiquitous in modern speech technologies. It is currently common practice to use ANNs from the second generation, as defined by Maass \cite{Maass1997}, in which the elementary units process real-valued inputs with an activation function and transmit real-valued outputs. This form of analogue communication actually differs from what is observed inside biological neural networks, where the information is encoded and transmitted in the form of sparse and binary sequences of events called spike trains. Activation functions in ANNs can be interpreted as firing rate approximations of an underlying spiking mechanism, but this neglects the precise spike timings, whose particular importance in the visual cortex and in auditory neurons has been demonstrated by several neuroscience studies  \cite{Mainen1995, Van2001, Butts2007, Gollisch2008}. 

Spiking Neural Networks (SNNs) are based on physiologically plausible neuron models \cite{Gerstner2002} that lead to more realistic neuronal dynamics and represent the third generation of ANNs \cite{Maass1997}. Whilst they can lead to energy-efficient hardware implementations, and have notably gained interest around keyword-spotting devices \cite{Blouw2019}, SNNs are typically harder to train and have not yet caught up with the ANN performance on speech processing tasks in general. Nevertheless, SNNs have recently achieved concrete progress on speech command recognition tasks \cite{Bittar2022a}. They are now able to compete with larger gated recurrent neural networks such as LSTMs \cite{Hochreiter1997} on the Google Speech Commands data set \cite{Warden2018}, which represents roughly 30 hours of speech data in the form of short command words. This recent success is mostly due to the surrogate gradient method \cite{Neftci2019}, that allows SNNs to be trained via gradient descent like ANNs, making them compatible with modern deep learning frameworks. There even exist CUDA implementations of the surrogate gradient backpropagation \cite{Shrestha2018, Bauer2022} that can speed up training on GPUs. 

Despite this apparent compatibility, their usage has so far been restricted to relatively simple tasks and small networks, when compared to current end-to-end ANN architectures. In particular, it appears that this method has not yet been thoroughly tested in an advanced sequence-to-sequence learning scenario. In this study, we take the method that has been shown to be successful on speech command recognition, and apply it to the more challenging task of large vocabulary continuous speech recognition (LVCSR). Even though the scalability of surrogate gradient SNNs to multi-layered architectures has been assessed \cite{Bittar2022a}, this was purely done within the constraints of rather simple tasks compared to LVCSR. Indeed, speech command recognition involves very short audio samples, for which the complete input sequence needs to be labelled with a single class. On the other hand, LVCSR must handle long speech utterances and predict the corresponding arrangements of tokens using encoder-decoder architectures that typically involve far more components and parameters. Wu et al. \cite{Wu2020} have used SNNs on LVCSR tasks, however, instead of the surrogate gradient approach, they use a tandem learning rule, that relies on sharing weights with non-spiking ANNs that approximate the spike counts, thereby discarding meaningful information about spike timings in the learning mechanism. The only piece of research involving surrogate gradient SNNs on LVCSR that we were able to find is the very recent work of Ponghiran and Roy \cite{Ponghiran2022}. Inspired by the LSTM, they define a custom version of SNNs that combines a forget gate with multi-bit outputs instead of binary spikes, and reach error rates of 19.72$\%$ and 11.75$\%$ on TIMIT \cite{timit} and LibriSpeech \cite{librispeech} (100 hours) respectively.

In this work, we come back to the standard most commonly used leaky integrate-and-fire (LIF) neuron model without gates or multi-bit outputs and show that it is already capable of replacing LSTM layers with only minor losses of performance on LVCSR tasks. Thereby we aim to,
\begin{enumerate}
    \item Assess the compatibility and scalability of standard surrogate gradient SNNs as part of end-to-end, sequence-to-sequence, deep architectures on LVCSR tasks.
    \item Test the representational capabilities of SNNs as speech encoders, in which the information is reduced to a sparse and binary form.
    \item Confirm that as opposed to recurrent ANNs, SNNs can be robust to exploding gradient problems without resorting to gates.
\end{enumerate}
More generally, our aim is not to improve upon the current ANN state-of-the-art results, but to assess the current capabilities of surrogate gradient SNNs on more advanced deep learning tasks. The results appear promising for the inclusion of SNNs into energy-efficient speech technologies.

\section{Single Neuron Model}
\label{sec:neuron}

\begin{figure}[t]
\centering
\includegraphics[width=0.9\columnwidth]{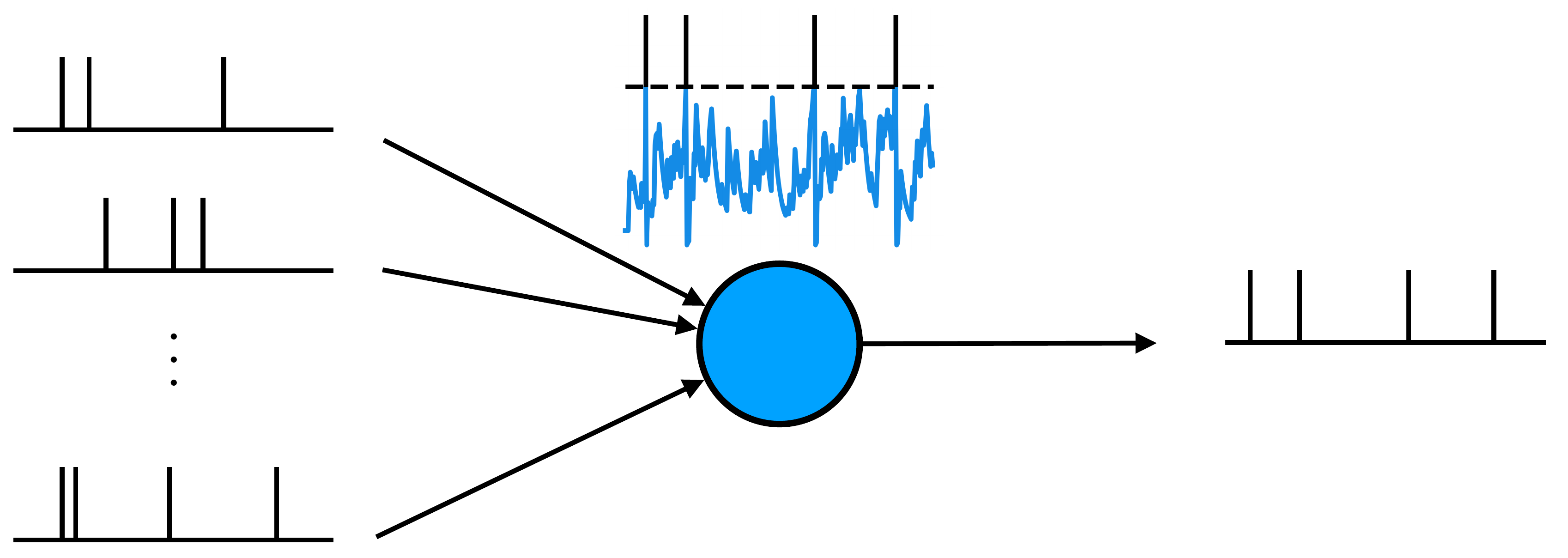}
\caption{Behaviour of a single spiking neuron that receives spike trains as inputs and produces an output spike train.}
\label{fig:neuron} 
\end{figure}

The behaviour of a single spiking neuron can be captured by the LIF neuron model, whose dynamics are characterised by four phases, (i) the integration of stimuli, (ii) a decay back to rest in the absence of stimuli, (iii) the emission of a short pulse when a critical threshold value is attained, and (iv) a recovery period after a spike is emitted. It can be implemented in discrete time as,
\begin{subequations}
\begin{align}
    \label{eq_It}
    I[t]&=\text{BN}\big(Wx[t]\big) + Vs[t-1] \\\label{eq_ut}
    u[t]&=\alpha\,\big(u[t-1]-s[t-1]\big)+\big(1-\alpha\big)I[t] \\\label{eq_st}
    s[t]&=\big(u[t]\geq \vartheta\big) \, .
\end{align}
\end{subequations}
Here the stimulus $I[t]$ is a linear combination of inputs $x[t]$ from feedforward connections $W$ and outputs $s[t-1]$ from recurrent connections $V$, where batch normalisation \cite{Ioffe2015} is applied to the former. The membrane potential $u[t]$ integrates the incoming stimuli, but also naturally decays back to its rest state at zero. This leaky behaviour is characterized by a trainable coefficient $\alpha\in[0,1]$. In order to reach its threshold value of $\vartheta=1$ and produce a spike $s[t]=1$, the neuron must therefore receive a critical amount of excitatory stimuli within a sufficiently short time window. After a spike is emitted, the membrane potential is reset to a lower value. Most of the time, the neuron therefore stays silent, i.e., $s[t]=0$, and no information is transmitted. These dynamics are illustrated in Figure \ref{fig:neuron}, where the membrane potential is shown in blue.

\noindent In an equivalent nonspiking recurrent neural network, the stimulus $I[t]$ is also computed as in Equation \eqref{eq_It}, but instead of using the spiking dynamics of Equations \eqref{eq_ut} and \eqref{eq_st}, an activation value is simply computed using a nonlinear function $g(\cdot)$, such as the rectified linear unit (ReLU) or sigmoid,
\begin{equation}
    y[t]=g\big(I[t]\big) \, .
\end{equation}
The real-valued outputs $y[t]$ can then be interpreted as mean spike counts or firing rates over some arbitrary period of time. Compared to such standard artificial neurons commonly used in deep learning, a spiking neuron has several particularities,
\begin{itemize}
    \item the transmission of sparse binary events instead of dense real-valued sequences, which can reduce the number of operations and the energy consumption.
    \item an additional parameter $\alpha$, responsible for how long the neuron will remember certain information.
    \item in principle a better robustness to exploding gradient problems due to the sparse and binary properties of the variable to which recurrence is applied.
\end{itemize}

\section{Surrogate Gradient Method}
\label{sec:surrogate}

In order to train spiking neural networks, one can either emulate biologically inspired learning rules, such as spike timing dependent plasticity (STDP) \cite{Dan2006}, or employ gradient based methods, which have recently proven to be extremely successful for training deep artificial neural networks. In this paper, with the objective of making SNNs compatible with modern deep learning frameworks, we concentrate on the ubiquitous stochastic gradient descent (SGD) method. This also allows training of SNNs and ANNs jointly as hybrid architectures inside the same framework.

As a result of the non-differentiable threshold operation defined in Equation \eqref{eq_st}, SNNs are not directly compatible with SGD. Nevertheless, as reviewed by Neftci et al. \cite{Neftci2019}, the surrogate gradient method allows replacement of the undefined gradient with a surrogate during the backward pass.  This in turn allows SNNs to be trained like recurrent neural networks (RNNs) using the Back-Propagation Through Time (BPTT) algorithm, which is a generalisation of gradient descent for handling sequential data. As presented in the introduction, this method has recently proven to be reliable on speech command recognition tasks \cite{Bittar2022a}, but its applicability to deeper networks with the more involved task of LVCSR remains to be assessed.

Different choices of functions have been tested to define the pseudo-derivative of the Heaviside step function \cite{Zenke2018, Shrestha2018, Bellec2018, Kaiser2020, Yin2021}. All share the characteristics of being positive when the potential $u[t]$ is close to the threshold value $\vartheta$ and vanishing otherwise. In this work, similarly to \cite{Kaiser2020, Bittar2022a}, we use the so-called boxcar function,
\begin{equation}
    \frac{\partial s[t]}{\partial u[t]}=
    \begin{cases}
    0.5 \quad&\text{if}\quad |u[t]-\vartheta|\leq 0.5 \\
    0 \quad&\text{otherwise,}
    \end{cases}
\end{equation}
which we manually set as the custom backward pass of Equation \eqref{eq_st} and exploit auto-differentiation inside the deep learning framework PyTorch \cite{Paszke2017}.

\section{Experiments}
\label{sec:exp}

Using the SpeechBrain framework \cite{speechbrain}, we perform experiments on three speech recognition tasks with increasing amounts of training data, TIMIT (4 hours), LibriSpeech100 (100 hours) and LibriSpeech960 (960 hours). In our baseline, we use LSTMs that have proven representational capabilities for speech. We have no hypothesis that SNNs can replace all of LSTMs capabilities. Nevertheless we would expect simpler recurrence to suffice for at least some of the layers.

\subsection{Baseline architecture}

On all three tasks, the inputs consist of 40 mel-filterbank features that are extracted from the utterance waveforms. They are passed through convolutional layers, where time pooling is applied before reaching four bidirectional recurrent LSTM layers of 512 hidden units each. Some additional linear layers output the desired phoneme or subword probabilities. On TIMIT, these probabilities, which correspond to 40 different phoneme labels (including a blank class), directly go into a connectionist temporal classifier (CTC) loss \cite{Graves2006}. On LibriSpeech (both 100 and 960 hours), the probabilities correspond to 5000 byte-pair encoding subword units \cite{Sennrich2015}. In addition to CTC loss, an attention-based RNN decoder is also used during training with Negative Log-Likelihood (NLL) loss applied to the predicted sequence. Finally, at inference time, a pretrained Transformer-based language model is used instead of the RNN decoder. 

\subsection{Gradually replacing LSTMs}

\begin{figure*}
\centering
\includegraphics[width=0.8\textwidth]{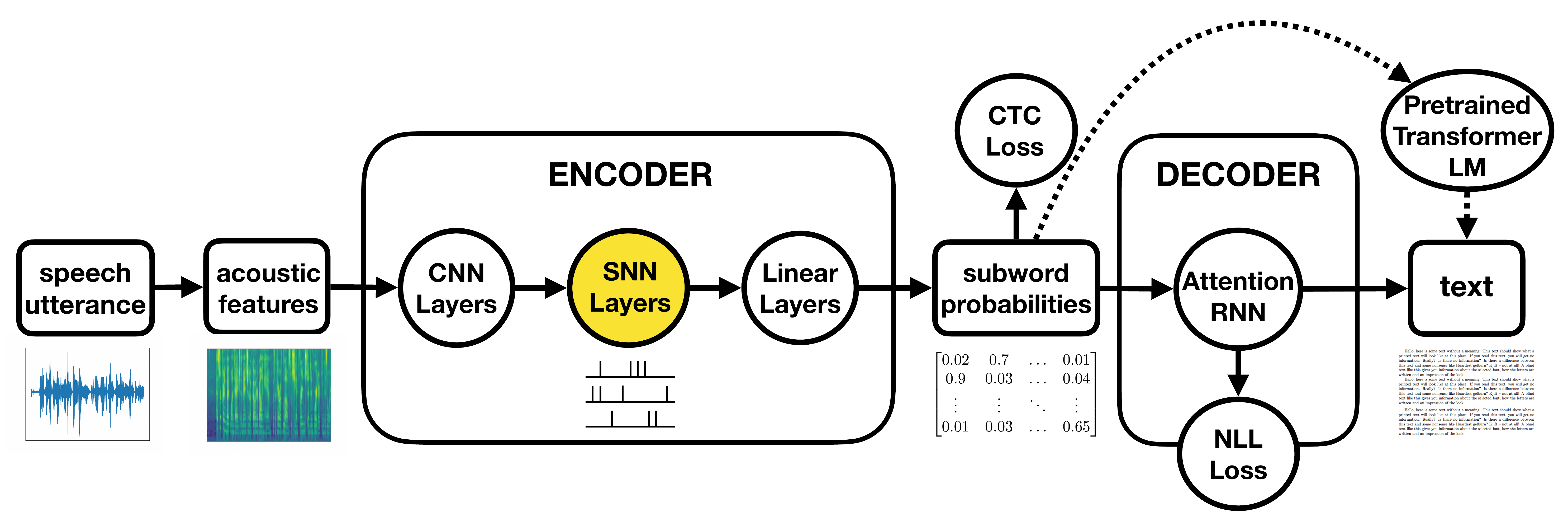}
\caption{Sequence-to-sequence LVCSR pipeline, where the LSTM layers from the SpeechBrain baseline have been replaced by SNNs inside the encoder. A pretrained Transformer-based language model is used at inference time for testing.}
\label{fig:asr} 
\end{figure*}

The LSTM layers inside the encoder are the only part of the end-to-end training pipeline that we focus on for our experiments. We gradually replace them with SNN layers defined using Equations \eqref{eq_It}-\eqref{eq_st}, and reach the architecture illustrated in Figure \ref{fig:asr}.
In order to compare with equivalent nonspiking layers, we also make similar experiments by gradually replacing the LSTM layers with standard non-gated RNN layers instead of the recurrent spiking layers. Note that both are referred to as RNNs in the tables. For instance, \emph{1 RNN - 3 LSTM} means that we replaced the first LSTM layer with a non-gated RNN and kept the three remaining ones as they were. For each RNN-LSTM configuration, we therefore train two instances of the architecture, once with spiking neurons and once with nonspiking neurons in the RNN layers.

\subsection{Credible intervals on error rates}

On TIMIT, the test set consists of 192 sentences that contain a total of $n=7193$ phonemes. On LibriSpeech, the clean test set, which is used for both 100h and 960h experiments, represents 2620 sentences for a total of $n=52576$ words. When testing an ASR model, the error rate corresponds to the fraction of words or phonemes that were incorrectly predicted by the model. The number of successes in such an experiment with binary outcomes can be modelled as a binomial distribution. For trivial priors, the posterior of the binomial distribution is beta distributed. Using the equal-tailed 95$\%$ credible intervals on the posterior distribution, we get error bars between $\pm0.84\%$ and $\pm0.88\%$ for the reported phoneme error rates (PERs) on TIMIT, and about $\pm0.21\%$ for the reported word error rates (WERs) on LibriSpeech.

\subsection{TIMIT experiments}

On TIMIT, as presented in Table \ref{table:timit}, we observe that gradually replacing the LSTM layers by non-gated recurrent spiking layers only hinders the accuracy by less than two percent. Similar results are obtained when replacing the LSTM layers by standard nonspiking RNNs instead. This first shows that surrogate gradient SNNs are capable of being trained as part of an end-to-end hybrid architecture. Moreover, they perform similarly to their nonspiking RNN counterpart, which suggests that sparse, binary information is sufficient to encode speech information. It is worth noting that even if LSTMs remain marginally better, they appear to be over-parameterized when used on such a small data set.

\begin{table}[h]
\centering
\caption{PERs on TIMIT test set.}
\label{table:timit}
\begin{tabular}{c c c} 
 \toprule
 \textbf{Encoder} &  \textbf{Spiking} & \textbf{Nonspiking}\\ [0.5ex] 
 \midrule
 0 RNN - 4 LSTM  & & 15.77$\%$ \\
 1 RNN - 3 LSTM  & 16.11$\%$ & 16.07$\%$ \\ 
 2 RNN - 2 LSTM  & 16.59$\%$ & 17.02$\%$\\ 
 3 RNN - 1 LSTM  & 16.78$\%$ & 16.75$\%$\\ 
 4 RNN - 0 LSTM  & 17.63$\%$ & 16.31$\%$ \\
 \bottomrule
 \end{tabular}
\end{table}

\begin{table}[h]
\centering
\caption{WERs on LibriSpeech100 clean test set.}
\label{table:libri100}
\begin{tabular}{c c c} 
 \toprule
 \textbf{Encoder} &  \textbf{Spiking} & \textbf{Nonspiking}\\ [0.5ex] 
 \midrule
 0 RNN - 4 LSTM  & & 5.91$\%$ \\
 1 RNN - 3 LSTM  & 6.31$\%$ & 5.99$\%$ \\ 
 2 RNN - 2 LSTM  & 6.66$\%$ & diverged \\ 
 3 RNN - 1 LSTM  & 7.13$\%$ & diverged \\ 
 4 RNN - 0 LSTM  & 10.56$\%$ & diverged \\
 \bottomrule
 \end{tabular}
\end{table}

\begin{table}[h]
\centering
\caption{WERs on LibriSpeech960 clean test set.}
\label{table:libri960}
\begin{tabular}{c c c} 
 \toprule
 \textbf{Encoder} &  \textbf{Spiking} & \textbf{Nonspiking}\\ [0.5ex] 
 \midrule
 0 RNN - 4 LSTM  & & 3.46$\%$ \\
 1 RNN - 3 LSTM  & 3.51$\%$ & diverged \\ 
 2 RNN - 2 LSTM  & 4.17$\%$ & diverged \\ 
 3 RNN - 1 LSTM  & 4.69$\%$ & diverged \\ 
 4 RNN - 0 LSTM  & 9.94$\%$ & diverged \\
 \bottomrule
 \end{tabular}
\end{table}

\subsection{LibriSpeech experiments}

The results on the 100h version are presented in Table \ref{table:libri100}. We first notice that replacing the first LSTM layer with either a spiking or a nonspiking RNN layer only marginally degrades the performance, even though a non-gated layer involves four times fewer trainable parameters compared to a LSTM layer. When replacing more than one LSTM layer, nonspiking RNNs suffered from exploding gradient issues, whereas equivalent spiking networks remained stable. This confirms that spiking RNNs are more robust to the problems of exploding gradients, compared to standard non-gated RNNs. We can assume that this is due to the sparse nature of the information that prevents gradients from excessively accumulating. On top of being able to learn where nonspiking RNNs diverged, SNNs were actually capable of replacing LSTM layers. The error rates using SNNs remain close to a 1$\%$ difference about the LSTM performance, except when replacing all LSTM layers. This suggests that the gating mechanism present in LSTMs may be important for an adequate processing of speech information, even if only one layer may suffice. It is worth noting that we also compared with fewer LSTM layers and ensured that adding SNNs was indeed beneficial. For instance, two SNN layers followed by two LSTM layers did outperform two LSTM layers on their own. These results were left out of the tables for clarity.

The results on the complete 960h training set presented in Table \ref{table:libri960} show the same tendency as with the 100h reduced version. Here nonspiking RNNs suffered from exploding gradients and were not able to replace a single LSTM layer, whereas spiking RNNs were even capable of replacing all of them. Again, keeping one LSTM layer significantly improves the performance, which supports our hypothesis that SNNs cannot replace all of the LSTMs capabilities.

\section{Conclusion}
\label{sec:conclusion}

In the introduction, we set three goals for our work on spiking neural networks. After successfully training them in the context of LVCSR, we can conclude that surrogate gradient SNNs are compatible with large end-to-end, sequence-to-sequence modern architectures. This answers our first and more general goal of assessing their scalability to more advanced tasks and deeper networks. Secondly, on all tasks, spiking layers, which involve four times fewer trainable parameters, were shown to be capable of replacing LSTMs with only minor losses in performance. Even though keeping a single LSTM layer still significantly helped reducing the error rate, this demonstrates that the information inside neural networks can efficiently be reduced to sparse and binary events without considerably affecting the network encoding capabilities. Thirdly, recurrent SNNs proved to be robust to exploding gradient problems without requiring gates, which was not possible using standard RNNs on large data sets. More generally, these findings contribute to making SNNs viable deep learning components and promising tools for energy-efficient speech technology.

\vfill\pagebreak
\section{References}
\ninept
\printbibliography[heading=none]

\end{document}